\documentclass[10pt,twocolumn,letterpaper]{article}

\usepackage{iccv}
\usepackage{times}
\usepackage{epsfig}
\usepackage{graphicx}
\usepackage{amsmath}
\usepackage{amssymb}

\usepackage{ctable}
\usepackage{array}
\usepackage{caption}
\usepackage{hhline}
\usepackage{multirow}

\newcolumntype{L}[1]{>{\raggedright\let\newline\\\arraybackslash\hspace{0pt}}m{#1}}
\newcolumntype{C}[1]{>{\centering\let\newline\\\arraybackslash\hspace{0pt}}m{#1}}
\newcolumntype{R}[1]{>{\raggedleft\let\newline\\\arraybackslash\hspace{0pt}}m{#1}}
\usepackage[breaklinks=true,bookmarks=false]{hyperref}

\iccvfinalcopy 

\def\iccvPaperID{2194} 

\ificcvfinal\fi
\begin{document}

\title{Holistic Planimetric prediction to Local Volumetric prediction for 3D Human Pose Estimation}

\author{Gyeongsik Moon\\
ASRI, Seoul National University\\
{\tt\small mks0601@snu.ac.kr}
\and
Ju Yong Chang\\
Kwangwoon University\\
{\tt\small juyong.chang@gmail.com}
\and
Yumin Suh\\
ASRI, Seoul National University\\
{\tt\small n12345@snu.ac.kr}
\and
Kyoung Mu Lee\\
ASRI, Seoul National University\\
{\tt\small kyoungmu@snu.ac.kr}
}

\maketitle

\begin{abstract}

We propose a novel approach to 3D human pose estimation from a single depth map. Recently, convolutional neural network (CNN) has become a powerful paradigm in computer vision. Many of computer vision tasks have benefited from CNNs, however, the conventional approach to directly regress 3D body joint locations from an image does not yield a noticeably improved performance. In contrast, we formulate the problem as estimating per-voxel likelihood of key body joints from a 3D occupancy grid. We argue that learning a mapping from volumetric input to volumetric output with 3D convolution consistently improves the accuracy when compared to learning a regression from depth map to 3D joint coordinates. We propose a two-stage approach to reduce the computational overhead caused by volumetric representation and 3D convolution: Holistic 2D prediction and Local 3D prediction. In the first stage, Planimetric Network (P-Net) estimates per-pixel likelihood for each body joint in the holistic 2D space. In the second stage, Volumetric Network (V-Net) estimates the per-voxel likelihood of each body joints in the local 3D space around the 2D estimations of the first stage, effectively reducing the computational cost. Our model outperforms existing methods by a large margin in publicly available datasets.

\end{abstract}


\section{Introduction}

Accurate 3D human pose estimation is an important requirement for activity recognition, and it has diverse applications such as in human computer interaction or augmented reality~\cite{romero2009monocular}. It has been studied for decades in the computer vision community and has been attracting considerable research interest again due to the introduction of low-cost depth cameras.

\begin{figure}[t]
\begin{center}
   \includegraphics[width=1.0\linewidth]{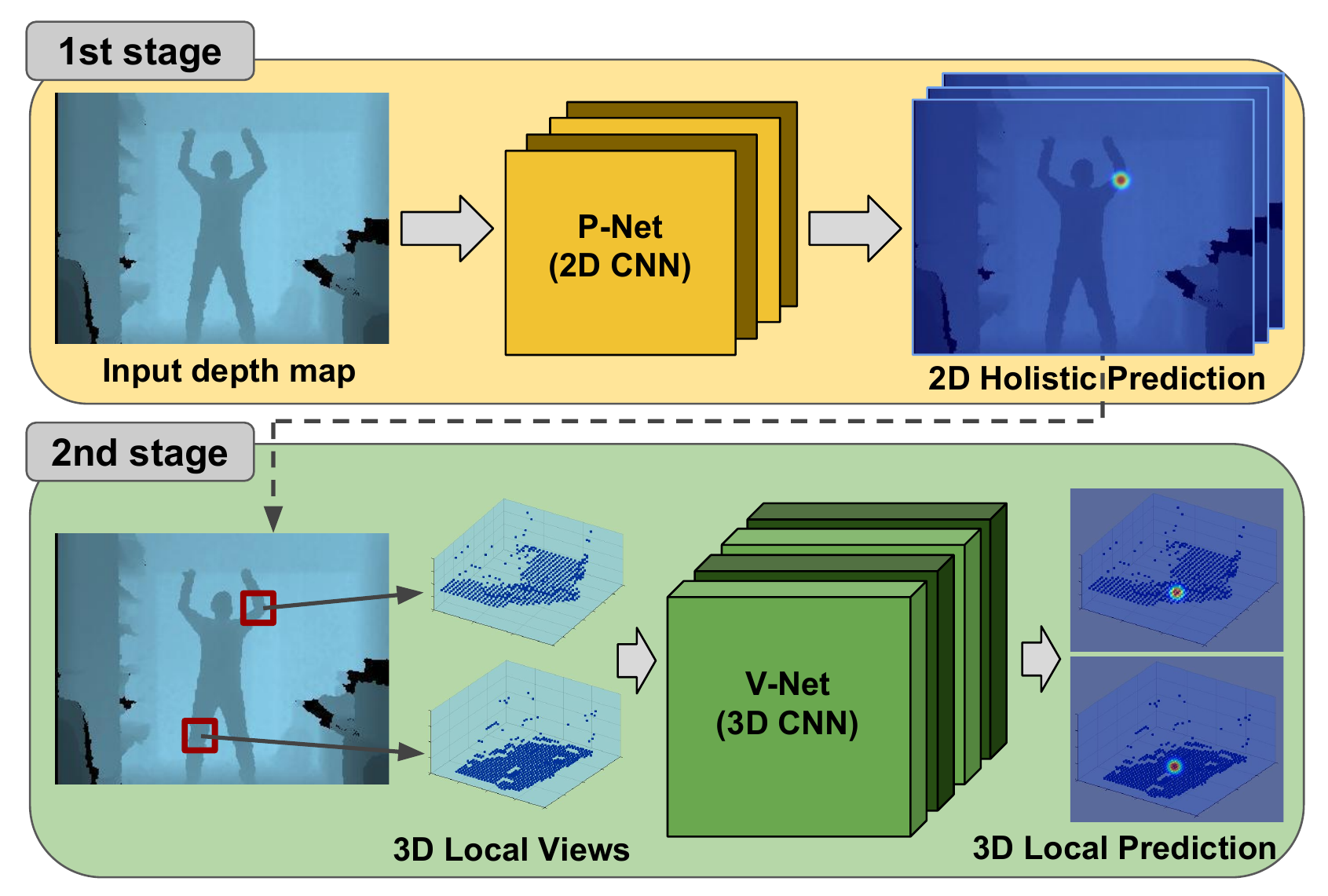}
\end{center}
   \caption{Illustration of the proposed system. In the first stage, P-Net estimates the per-pixel likelihood for each body joint in the holistic 2D space. On the basis of the 2D estimation, 3D local views are generated for each body joint in the form of an occupancy grid~\cite{maturana2015voxnet}, which becomes an input to the second stage. In the second stage, V-Net takes the 3D local views as an input and performs 3D local prediction where the per-voxel likelihood in the 3D local space is estimated, which subsequently gives accurate 3D body joint positions.}
\label{fig:brief_architecture}
\end{figure}

Recently, powerful discriminative approaches such as convolutional neural network (CNN) are outperforming existing methods in various computer vision tasks such as object detection~\cite{ren2015faster,liu2016ssd}, semantic segmentation~\cite{long2015fully}, and 2D human pose estimation~\cite{newell2016stacked,wei2016convolutional}. However, CNN does not yield a noticeably improved performance in 3D pose estimation despite its richer information~\cite{haque2016towards} because of  the highly non-linear relations between convolutional feature maps and Cartesian coordinates. It is known that this non-linear module converting the 2D discretized CNN features into the 3D continuous coordinates makes network suffer from inaccurate estimation~\cite{tompson2014joint}. To resolve the non-linearity problem in 2D human pose estimation, Tompson \etal~\cite{tompson2014joint} proposed to estimate per-pixel likelihood rather than directly regress the 2D joint coordinates. Thus, it overcomes the problem of the existing coordinate regression methods, and most of the recent works~\cite{newell2016stacked,wei2016convolutional,bulat2016human} utilize the per-pixel likelihood to estimate the body joint positions. Therefore, estimating per-voxel likelihood for the 3D human pose estimation is a natural solution. However, the process requires 3D tensor representation and 3D convolution, which significantly increase the computational cost of the entire system.

\begin{figure}[t]
\begin{center}
\includegraphics[width=1.0\linewidth]{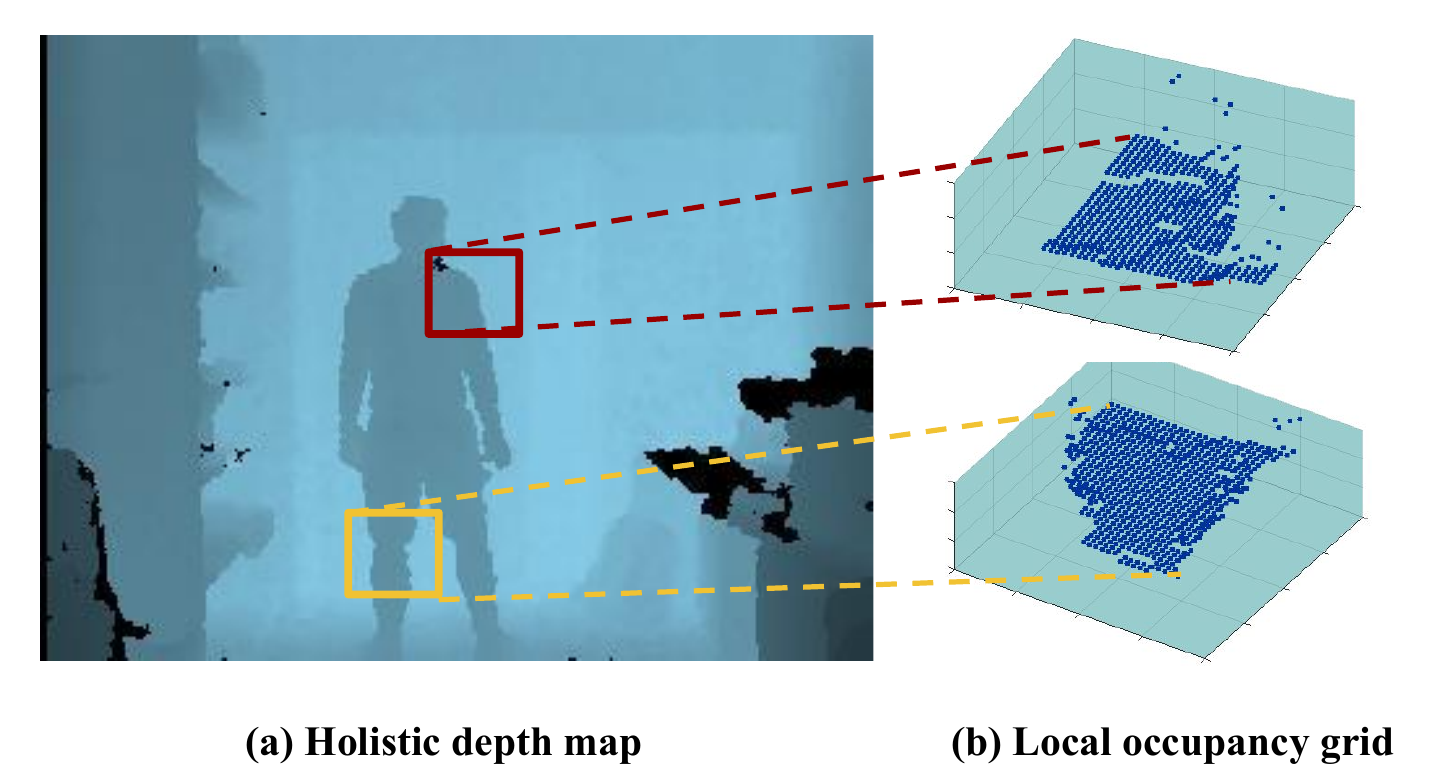}
\end{center}
   \caption{Visualization of the 3D occupancy grid around the left shoulder and the right knee. From the depth map (a), we crop local patches around the body joints estimated from the P-Net and construct the corresponding local occupancy grids as in (b).}

\label{fig:occupancy_grid}
\end{figure}

In this paper, we propose a novel 3D human pose estimation approach based on the following ideas: First, we represent the input depth map in a form of 3D occupancy grid following ~\cite{maturana2015voxnet} as in Figure~\ref{fig:occupancy_grid}. Then we feed it into 3D CNN to estimate per-voxel likelihood for body joints localization instead of regressing a set of joint coordinates directly. We train our network by supervising the per-voxel likelihood for each body joint, which provides richer information than 3D coordinates. We show that estimating per-voxel likelihood from the occupancy grid using 3D CNN significantly improves the performance compared with the traditional coordinate regression methods based on 2D representations. Second, to resolve the increased computational cost caused by 3D representations (i.e., the occupancy grid), we separate 3D human pose estimation into two stages, namely, holistic 2D prediction and local 3D prediction as in Figure~\ref{fig:brief_architecture}. In the first stage, {\it Planimetric Network} (P-Net), a 2D CNN, estimates the per-pixel likelihood for each body joint from a single depth map. On the basis of the P-Net estimation, we generate 3D local views for each body joint in the form of the occupancy grid, which is then used as an input to the second stage. In the second stage, the 3D local views are fed into {\it Volumetric Network} (V-Net), a 3D CNN, to estimate per-voxel likelihood for each body joint in the local 3D space. We can reduce the computational overhead by considering the 3D local space that can be obtained from the P-Net, which accurately estimates the 2D location for each joint. Our system outperforms existing methods by a large margin in publicly available datasets.

We summarize our contributions as follows: 

\begin{itemize}
\item We introduce a novel 3D human pose estimation approach from a single depth map. The core idea of our approach is the per-voxel likelihood estimation from the 3D occupancy grids instead of the Cartesian coordinate regression from the 2D representation of the input depth map.

\item We formulate the 3D human pose estimation problem as a two-stage approach: holistic 2D prediction followed by local 3D prediction, which can mitigate the computational overhead caused by 3D convolution and 3D occupancy grid representation.

\item We conduct thorough experiments using various real datasets, which shows that the proposed method produces significantly more accurate results than the state-of-the-art methods in publicly available datasets.
\end{itemize}

\section{Related Work}
{\bf Depth-based 3D human pose estimation.} 
Generative and discriminative models for depth-based 3D human pose estimation exist. The generative models estimate the pose by finding the correspondences between the pre-defined body model and the input 3D point cloud. The iterative closest point algorithm is commonly used for 3D body tracking~\cite{ganapathi2012real,grest2005nonlinear,knoop2006sensor,helten2013personalization}. Recently, Ye \etal~\cite{ye2014real} formulated a human body as a Gaussian mixture model. By contrast, the discriminative models do not require the body template, and they directly estimate the positions of body joint. Shotton \etal~\cite{shotton2013real} trained random forests to classify each pixel into one of the pre-defined body parts. The final coordinates of body joints are obtained by the mean-shift mode detection algorithm. Also, Girshick \etal~\cite{girshick2011efficient} used random forests to directly regress the coordinates of body joints from the input depth map. Their method generates a 3D point cloud for each joint, which is then clustered by the mean-shift algorithm. Jung \etal~\cite{jung2016sequential} used a similar method; however, they separated the 3D pose estimation problem into the localization and identification stages to obtain a more accurate estimation. Jung \etal~\cite{yub2015random} used a random tree walk algorithm, which can reduce running time significantly. Recently, Haque \etal~\cite{haque2016towards} proposed the viewpoint invariant pose estimation method using CNN and multiple rounds of recurrent neural network. Their model learns viewpoint-invariant features, which make the model robust to viewpoint variation. Also, their network iteratively refines the estimation of the previous round to estimate a more accurate pose. However, they estimated the set of 3D coordinates to localize body joints through highly non-linear mapping.

\begin{figure}[t]
\begin{center}
\includegraphics[width=1.0\linewidth]{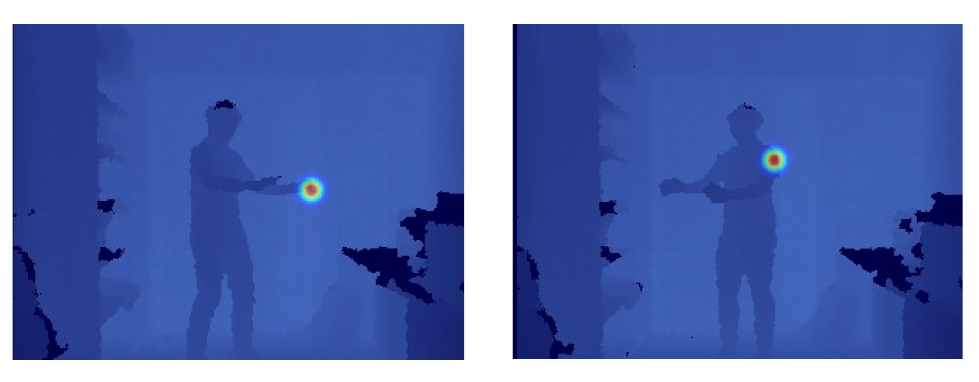}
\end{center}
   \caption{Visualization of the P-Net estimation. The P-Net estimates the per-pixel likelihood for each body joint. The left figure shows the likelihood for the left hand, and the right figure shows that for the left shoulder.}

\label{fig:2d_result}
\end{figure}

{\bf 2D human pose estimation using CNN.}
Recently, 2D human pose estimation research has shifted from classical approaches based on hand-crafted features~\cite{bourdev2009poselets,felzenszwalb2008discriminatively,yang2013articulated} to CNN-based approaches~\cite{tompson2014joint,newell2016stacked,wei2016convolutional,bulat2016human}. Toshev \etal~\cite{toshev2014deeppose} directly estimated the Cartesian coordinates of body joints using the multi-stage deep network and obtained a state-of-the-art performance. Tompson \etal~\cite{tompson2014joint} estimated the per-pixel likelihood for each joint using CNN and used it as the unary term for the external graphical model to accurately estimate joint positions. They argued that the direct regression approach in Toshev \etal~\cite{toshev2014deeppose} suffers from the highly non-linear mapping which causes inaccurate performance in high-precision regions. Following  Tompson \etal~\cite{tompson2014joint}, most of the recent CNN-based methods estimate the per-pixel likelihood for each body joint. Liu \etal~\cite{wei2016convolutional} used multiple stages of refinement to enlarge receptive fields. Newell \etal~\cite{newell2016stacked} used hourglass structure to exploit information from multiple scales and achieved state-of-the-art performance in a publicly available benchmark. Bulat and Tzimiropoulos~\cite{bulat2016human} used detection subnetwork to help regression subnetwork accurately localize body joint. Our proposed method follows the trend in 2D human pose estimation and predicts the per-voxel likelihood for each body joint instead of the 3D coordinates.

{\bf Volumetric representation and 3D CNN.}
Wu \etal~\cite{wu20153d} introduced volumetric representation of depth image and surpassed the existing hand-crafted descriptor based methods in 3D shape classification and retrieval. They represent each voxel as binary random variables and used convolutional deep belief network to learn the probability distribution for each voxel. Several recent works~\cite{maturana2015voxnet,song2016deep} also represent 3D input data as a volumetric form for 3D object classification and detection. Our work follows the strategy from ~\cite{maturana2015voxnet}. Maturana and Scherer~\cite{maturana2015voxnet} proposed several types of volumetric representation, occupancy grids, to fully utilize rich source of 3D information and efficiently dealing with large amount of point cloud data. They showed that their proposed CNN architecture and occupancy grids outperform those of Wu \etal~\cite{wu20153d}.

\section{Model overview}
The goal of our model is to estimate the 3D coordinates for all body joints. In the first stage, the P-Net takes a single depth map as an input and predicts the per-pixel likelihood for each body joint. On the basis of this estimation, the local patch of each body joint is cropped and converted to the 3D occupancy grids, which are fed into the second stage of our system. In the second stage, V-Net predicts the per-voxel likelihood in the 3D local space, and we can obtain the 3D position of each body joint by finding the maximum response. The overview of the proposed approach is illustrated in Figure~\ref{fig:brief_architecture}. We now describe two stages (i.e., P-Net and V-Net) of the proposed approach in the following sections.

\section{Holistic 2D Prediction}
\subsection{Model}
In the first stage, the proposed P-Net takes a single depth map, which includes a person, and estimates the per-pixel likelihood for each body joint, as shown in Figure~\ref{fig:2d_result}. Recently, many types of CNN-based approaches~\cite{tompson2014joint,newell2016stacked,wei2016convolutional,bulat2016human} that regress the per-pixel likelihood from an RGB image for 2D human pose estimation have been introduced; we employ the residual regression subnetwork~\cite{bulat2016human}. This architecture is a slightly modified version of the stacked hourglass network~\cite{newell2016stacked}, which exploits residual learning~\cite{he2016deep} with the information from multiple scales. The size of a human body joint is small in usual cases; thus, exploiting only the last convolutional feature maps to estimate the body joint position would hamper accurate estimation. Bulat and Tzimiropoulos~\cite{bulat2016human} used the residual skip connections across convolutional feature maps to exploit small-scale information. The input to our system is a depth map, not an RGB image; thus, we change the number of channels in the first layer of the network from three to one. We call the first stage holistic 2D prediction because the output space of the P-Net is the holistic view of body structure in 2D space. 

\begin{figure*}
\begin{center}
\includegraphics[width=1.0\linewidth]{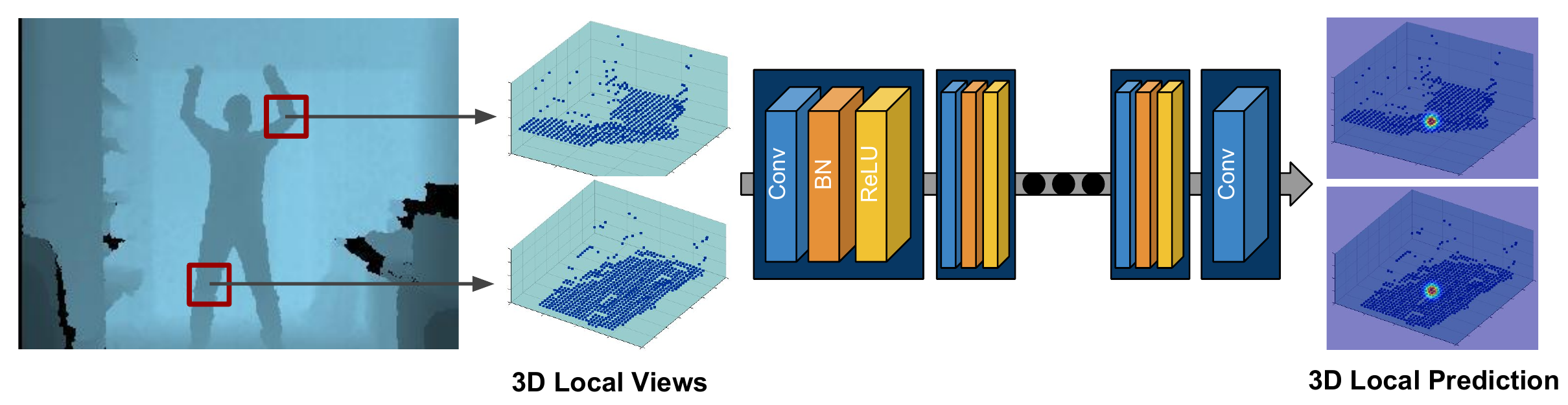}
\end{center}
   \caption{The architecture of the V-Net. In the second stage, the 3D local view is fed into the V-Net, which estimates the 3D local per-voxel likelihood for each body joint. All operations, such as convolution, batch normalization and pooling, are volumetric operations which means that the kernel has a shape of ($k_x$,$k_y$,$k_z$) instead of ($k_x$,$k_y$).}
\label{fig:second_architecture}
\end{figure*}

\subsection{Loss function}
To generate the ground truth heatmap (i.e., pixel-likelihood), we use a Gaussian peak whose mean is the joint position with \(\sigma\) = 5 pixels. We adopt the mean square error (MSE) as the loss function of the P-Net as follows:

\begin{equation}
L _{\mathrm{P}}= \frac{1}{J}\sum_{j=1}^{J} \sum_{x,y} \|h^j_*(x,y)-h^j(x,y)\|^2
\end{equation}
where $h_*$ is the ground truth heatmap, $h$ is the estimated heatmap for each body joint, and $J$ denotes the number of body joints.

\section{Local 3D Prediction}
\subsection{Occupancy grid model}
The occupancy grid model~\cite{moravec1985high,thrun2003learning} represents a 3D data as a 3D lattice of random variables that correspond to each voxel. The random variables have the probability of occupancy, which can be formulated as a function of input sensor data and prior knowledge. Maturana and Scherer~\cite{maturana2015voxnet} proposed several types of occupancy grid to perform 3D object detection from LiDAR and RGBD point clouds.

We follow the strategy of the {\it Hit grid} from ~\cite{maturana2015voxnet} to generate occupancy grid from a depth map, which does not distinguish free and unknown spaces. For a given depth patch, corresponding occupancy grid is represented as a 3D binary tensor $t$, where $t(x,y,z)=1$ if discretized depth at $(x,y)$ is $z$ and -1 otherwise. This representation can use the geometric information in the data better than when treating the depth map as an 2D image.

\subsection{Local prediction}
In the second stage, V-Net takes 3D local views of each body joint in the form of occupancy grid, which is generated from P-Net estimation and predicts body joint positions in the 3D local space. By using the 3D local views as input, we can exploit the accurate 2D estimation result from the first stage to reduce the search space of the second stage from the entire image to the cropped image. In addition, the small size representation enables the network to stack sufficient layers to enlarge the size of the receptive field. However, the size of the local view should be sufficiently large to fully capture the global context information of the input. Considering the trade-off between small and large input size, we set the size of the 3D local view (i.e., $x$$\times$$y$$\times$$depth$) as 32$\times$32$\times$40. Figure~\ref{fig:second_architecture} and Table~\ref{table:layer_specification} show the architecture and detailed configuration of the V-Net, respectively. The network contains 11 3D building blocks where each block consists of a 3D convolution layer and a 3D batch normalization layer followed by the non-linear activation layer (i.e., ReLU function). 

\begin{table}[]
\centering
\setlength\tabcolsep{1.0pt}
\def\arraystretch{1.3}
\begin{tabular}{C{1.0cm}||C{3.0cm}C{2.0cm}C{2.0cm}}
\specialrule{.1em}{.05em}{.05em} 
\#     &  \textbf{Layer}   & \textbf{Channel} & \textbf{Filter size} \\ \hline
1      &  Conv+BN+ReLU  &   64        &  7$\times$7$\times$7   \\ 
2      &  Conv+BN+ReLU   &   64        &  5$\times$5$\times$5       \\ 
3     &  Conv+BN+ReLU    &   128        &  5$\times$5$\times$5     \\ 
4    &  Conv+BN+ReLU     &   128        &  5$\times$5$\times$5       \\ 
5     &  Conv+BN+ReLU    &   128        &  5$\times$5$\times$5       \\ 
6     &  Conv+BN+ReLU     &   256        &  5$\times$5$\times$5      \\
7     &  Conv+BN+ReLU     &   256        &  5$\times$5$\times$5        \\
8      &  Conv+BN+ReLU   &   256        &  5$\times$5$\times$5       \\
9      &  Conv+BN+ReLU   &   256        &  1$\times$1$\times$1        \\
10     &  Conv+BN+ReLU    &   256        &  1$\times$1$\times$1       \\
11      &  Conv  &   \# of joints        &  1$\times$1$\times$1   \\ \specialrule{.1em}{.05em}{.05em} 
\end{tabular}
\caption{Layer specification of the V-Net. All convolutional layers have strides of 1. Note that network weights are three-dimensional arrays in $x$, $y$, $depth$ axes.}
\label{table:layer_specification}
\end{table}

\subsection{Per-voxel likelihood estimation}
We estimate the per-voxel likelihood to obtain the position of each body joint. It avoids regressing pose vectors directly from an image, which is a highly non-linear problem. As a result, our network does not need to output the unique position of each body joint but estimates the confidence in a discretized 3D tensor. This conversion from regressing pose vector to estimating confidence was previously proven beneficial in 2D human pose estimation~\cite{tompson2014joint}. Also, compared with the traditional highly non-linear method~\cite{haque2016towards}, which regresses the pose vector through multiple rounds of fully connected layers, our model has much fewer parameters and is translation-invariant because we do not use the fully-connected layers.

\subsection{Loss function}
To generate the ground truth heatmap (i.e., the per-voxel likelihood), we use a Gaussian peak whose mean is the joint position with \(\sigma\) = 1 pixel for all axes. We adopt the MSE as a loss function as follows:

\begin{equation}
L_{\mathrm{V}} = \frac{1}{J}\sum_{j=1}^{J} \sum_{x,y,z} \|h^j_*(x,y,z)-h^j(x,y,z)\|^2
\end{equation}
where $h_*$ is the ground truth heatmap, and $h$ is a estimated heatmap for each body joint. $J$ denotes the number of body joint, and $z$ denotes the discretized depth coordinate.

\section{Training and Optimization}
We train our two networks separately from scratch without using pre-trained models. All weights are initialized from a Gaussian distribution with zero-mean and \(\sigma\) = 0.001. For each model, gradient vectors are calculated from the loss function, and the weight is updated by the standard gradient descent algorithm with a mini-batch size of 4. The initial learning rate, weight decay, and momentum are set to $1\times10^{-4}$, $5\times10^{-4}$ and $9\times10^{-1}$, respectively. Both models are trained at 20K iterations with the initial learning rate and 10K iterations with learning rate divided by 10. The input of the P-Net is the randomly cropped 256$\times$256 patch from the 288$\times$288 input depth map for data augmentation, and pixel values are normalized between [-1, 1]. To train the V-Net, 32$\times$32$\times$40 occupancy grids are generated from the 2D estimation of the P-Net, and 32$\times$32$\times$36 grids are randomly cropped for data augmentation. We also attempted to train the V-Net based on the ground truth 2D position with up to 5 pixels of variation in the $xy$-space and tested based on the 2D estimation from the P-Net. However, this approach obtained a lower accuracy than that obtained by using the P-Net estimation result to train the V-Net. We believe it is because the distribution of the P-Net estimation is different from the 2D ground truth with the random $xy$-translation. Our model is implemented by Torch7~\cite{collobert2011torch7} and the NVIDIA TitanX GPU is used for training and testing.

\section{Experiments}

\subsection{Datasets}
{\bf Stanford EVAL dataset.}
The EVAL dataset~\cite{ganapathi2012real} consists of 9K front-view depth images. The dataset is recorded by a Microsoft Kinect camera at 30 fps and contains three actors performing eight sequences each. The ground truth of this dataset is the 3D coordinates of 12 body joints. Considering that this dataset does not distinguish between training and test sets, we used the leave-one-out strategy following ~\cite{yub2015random,haque2016towards}. The sequences from a person is used as the test set and the others are used as the training set. This procedure is repeated three times for each actor, and mean average precision is calculated to measure final performance.

{\bf ITOP dataset.}
The ITOP dataset is newly released in ~\cite{haque2016towards}. Compared with the stanford EVAL dataset, this dataset contains a large number of frames, consisting of 100K front-view and top-view depth images with 20 actors performing 15 sequences each. The ground truth of this dataset is the 3D coordinates of 15 body joints. As they published this dataset for viewpoint-invariant pose estimation, we use only front-view depth images for training and evaluation.

\subsection{Evaluation metrics}
{\bf Percentage of correct keypoints.}
To evaluate the performance of the P-Net, we use the percentage of correct keypoints (PCKh) as the evaluation metric. This metric proposed in ~\cite{andriluka20142d} classifies the estimated joint position as correct if the predicted joint is within 50\% of the head segment length to the ground truth joint.

\begin{figure}[t]
\begin{center}
   \includegraphics[width=1.0\linewidth]{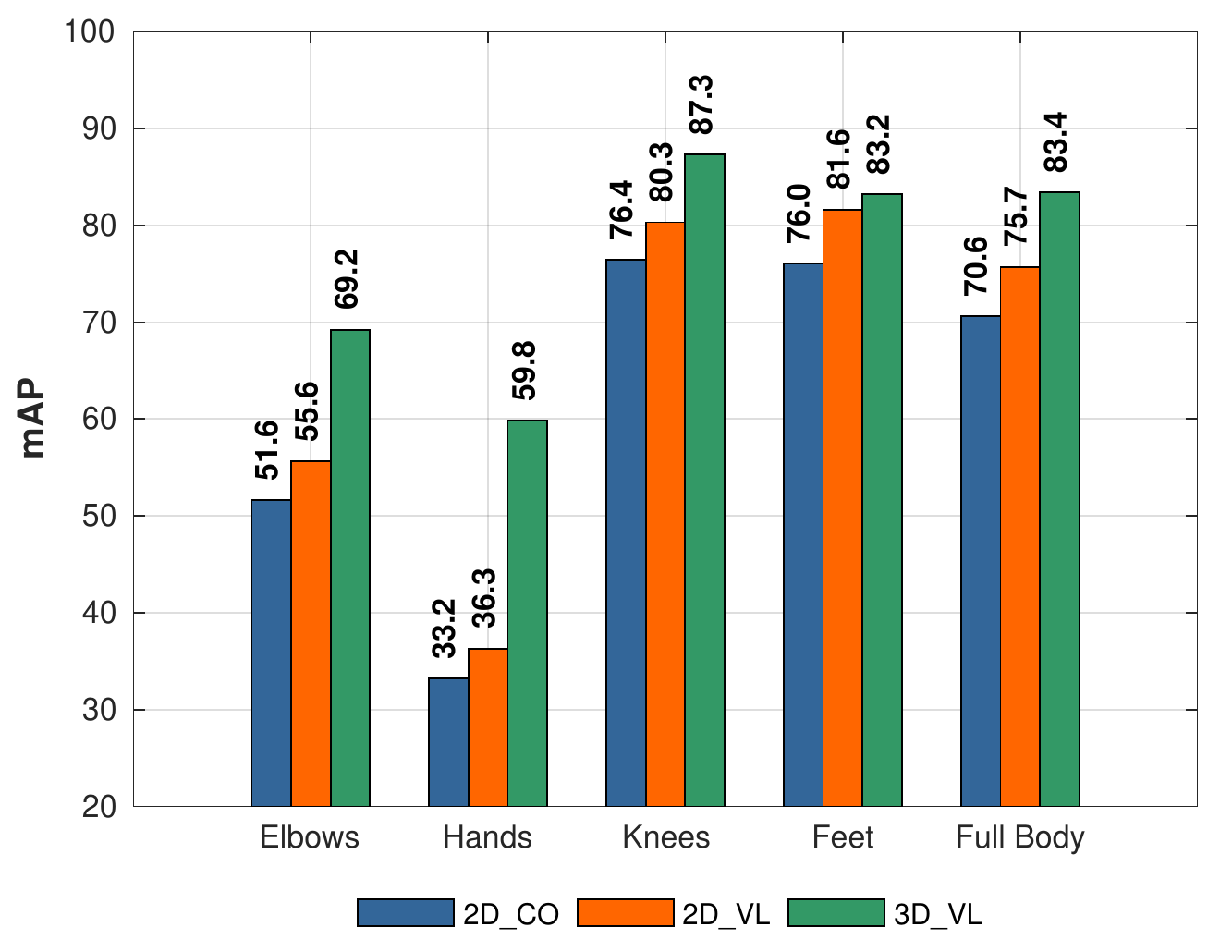}
\end{center}
   \caption{Performance according to the types of the input and output. {\it 2D\_CO} estimates the 3D coordinates from the local patch of the 2D depth map. {\it 2D\_VL} estimates the per-voxel likelihood from the same input. {\it 3D\_VL} estimates the per-voxel likelihood from the 3D occupancy grid. Only the hardest body joints are shown to avoid clutter.}
\label{fig:comparison_input_output}
\end{figure}

{\bf 10 cm rule.}
We use this metric to evaluate the performance of the proposed model following ~\cite{haque2016towards,yub2015random,shotton2013real}. This metric classifies the prediction of a model as correct when its estimated point and the ground truth point are within an error of 10 cm. Also, we use the mean average precision (mAP) to measure the performance of the model, which is the mean of average precisions of all body joints. For each body joint, the precisions are used to measure the performance. 

\subsection{Component Analysis}
We use the ITOP dataset to evaluate each component of our approach because this dataset has a larger amount of frames compared with the EVAL dataset.

\begin{table}[]
\centering
\setlength\tabcolsep{1.0pt}
\def\arraystretch{1.1}
\begin{tabular}{L{2.5cm}||C{1.1cm}C{0.9cm}C{0.9cm}C{0.9cm}C{1.5cm}}
\specialrule{.1em}{.05em}{.05em} 
   Input type  &  Elbows  & Hands & Knees & Feet & Full Body \\ \hline
3D local view      & \textbf{69.2}  &   \textbf{59.8}   &  \textbf{87.3} & \textbf{83.2} & \textbf{83.4} \\ 
3D holistic view      &  54.2  &   43.7  &  61.4  & 54.8 & 57.3 \\ \specialrule{.1em}{.05em}{.05em} 
\end{tabular}
\caption{Comparison between 3D local view and 3D holistic view input systems. Only the hardest body joints are shown to avoid clutter.}
\label{table:comparison_local_holistic}
\end{table}

{\bf 3D representation and per-voxel likelihood estimation.}
The main differences between traditional 3D pose estimation methods and our method is in the representation of input and output. We propose to use the 3D occupancy grid and per-voxel likelihood instead of the input 2D depth image and output 3D coordinates, respectively. To show the effectiveness of our proposed representation, we conduct an experiment, and its result is shown in Figure~\ref{fig:comparison_input_output}. The {\it 2D\_CO} estimates the 3D coordinates from the local patch of the 2D depth map and the {\it 2D\_VL} estimates the per-voxel likelihood from the local patch of the 2D depth map. The {\it 3D\_VL} is our proposed model. To implement the {\it 2D\_CO} and {\it 2D\_VL}, we converted all layers in Table~\ref{table:layer_specification} to the corresponding 2D layers (i.e., the shape of the kernel is ($k_x$,$k_y$) instead of ($k_x$,$k_y$,$k_z$)). In addition, we changed the last three building blocks into the fully-connected layers for coordinate estimation in the {\it 2D\_CO}. For the {\it 2D\_VL}, we replaced the last three layers to the convolutional layers whose output dimension is $J$$\times$$D$ following Pavlakos \etal~\cite{pavlakos2016coarse}, where $J$ is the number of joints, and $D$ is the discretized size of the depth axis, which becomes 36 as in the proposed method. The batch normalization and ReLU layers are followed by each newly added convolutional layer except the last layer. As the {\it 2D\_CO} and {\it 2D\_VL} of the figure show, estimating the per-voxel likelihood obtains better estimation results than regressing the 3D coordinates from a 2D image because the highly non-linear mapping between convolutional feature maps and 3D coordinates becomes problematic in the learning procedure. Also, as the {\it 2D\_VL} and {\it 3D\_VL} of the figure show, representing input with the 3D occupancy grids increases the accuracy, which means that our occupancy grid representation can fully utilize geometric information than simply treating the depth map as a 2D image. When comparing the {\it 2D\_CO} and the proposed model (i.e., {\it 3D\_VL}), our model requires 15 times fewer parameters while achieving 12.8 point higher accuracy. Note that we cannot test the combination of 3D input occupancy grid and 3D output coordinates because the required number of parameters is so large that the model does not fit into current available GPUs. 

\begin{figure}[t]
\begin{center}
   \includegraphics[width=1.0\linewidth]{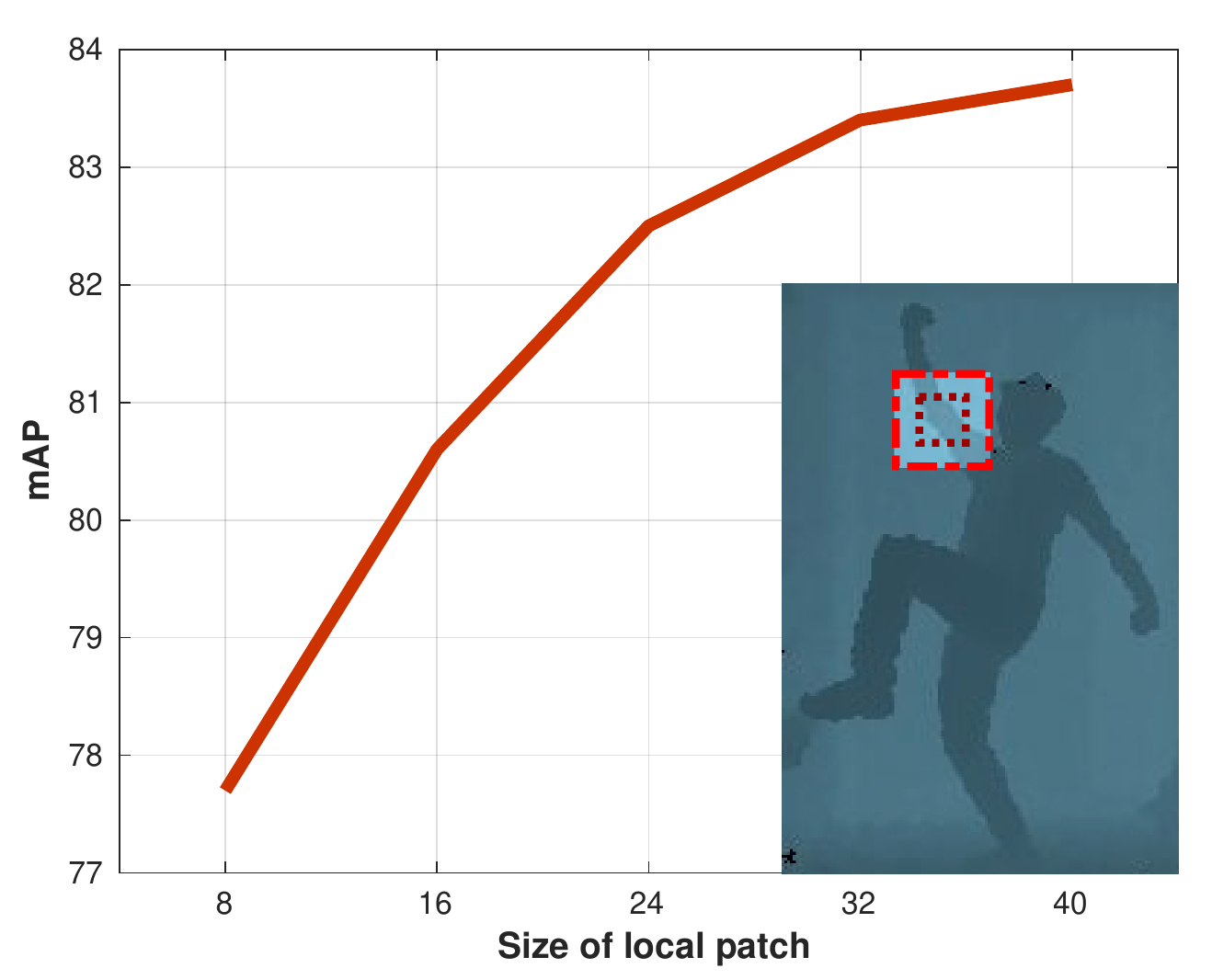}
\end{center}
   \caption{mAP changes with the increase in local patch size. The visualized human shows that as the local patch of the right elbow becomes larger, information from a neighboring body joint such as right shoulder can be captured, thereby providing evidence of the location of the right elbow.}
\label{fig:localSz}
\end{figure}

\begin{table}[]
\centering
\setlength\tabcolsep{1.0pt}
\def\arraystretch{1.1}
\begin{tabular}{L{2.0cm}||C{1.45cm}C{1.45cm}C{1.45cm}|C{1.45cm}}
\specialrule{.1em}{.05em}{.05em} 
     Body Part  &~\cite{yub2015random} &~\cite{haque2016towards} & Ours & Ours$^*$ \\ \hline
    Head   & 90.9 & \textbf{93.9} &   91.6 & 96.2 \\
    Neck  & 87.4 & \textbf{94.7} &   93.8 & 97.8\\
    Shoulders  & 87.8 & 87.0 &   \textbf{89.0} & 96.7 \\
    Elbows    & 27.5 & 45.5 &  \textbf{69.6} & 92.2  \\
    Hands   & 32.3 & 39.6 &  \textbf{69.0}  & 89.8 \\ \hhline{-----}
   Knees     & 83.4 & \textbf{86.0} &  68.7  & 97.1 \\
    Feet   & 90.0 & \textbf{92.3} &  89.7  & 94.0 \\ \hhline{=====}
    Upper Body    & 59.2 & 73.8 & \textbf{80.1} & 93.9 \\
  Lower Body      & 86.7 & \textbf{89.2} &  79.2& 95.5 \\
     Full Body  & 68.3 & {\color{blue} \textbf{74.1}} &   {\color{red} \textbf{79.8}}& 94.5 \\ \specialrule{.1em}{.05em}{.05em} 
\end{tabular}
\caption{Comparison with the existing methods on the EVAL dataset. {\it Ours$^*$} is the performance of the proposed method trained and tested from the 2D ground truth positions. The proposed model outperforms the existing methods.}
\label{table:comparison_EVAL_STOA}
\end{table}

{\bf Local views.}
The second stage of our system takes the local views as an input and locally estimates the positions of the body joint. Given that the local view is represented as a small size of tensor, the V-Net can increase the size of the receptive field to the input size by using sufficiently stacked layers. However, as it contains only local part of a human, it cannot fully exploit the global context information. By contrast, the holistic view can exploit contextual information better than the local view. However, because of an increased dimensionality due to the large size of the 3D holistic view, the network cannot stack enough layers, thereby resulting in the small size of the receptive field. Taking into account the increase in size of the receptive field is an important factor for human pose estimation~\cite{wei2016convolutional,newell2016stacked}, small size of the receptive field can hamper the accurate localization of body joints.
To analyze the trade-off between the local view input and the holistic view input, we build a model whose input is a 3D holistic view and estimates per-voxel likelihood in 3D global space. To generate the input of the newly built model, we cropped the human by utilizing the P-Net estimation to reduce the computational complexity and memory waste. The cropped human bounding box is resized to 128$\times$128 and converted to the 128$\times$128$\times$40 occupancy grid. We use the same data augmentation strategy as the proposed model. The size of the output tensor (i.e., 32$\times$32$\times$9) becomes smaller than that of the input tensor (i.e., 128$\times$128$\times$36) due to two max-pooling layers. The comparison result is shown in Table~\ref{table:comparison_local_holistic}. As the result shows, our local prediction localizes body joint much more accurately because of sufficiently stacked layers that enlarge the size of the receptive field.

\begin{table}[]
\centering
\setlength\tabcolsep{1.0pt}
\def\arraystretch{1.1}
\begin{tabular}{L{1.8cm}||C{0.8cm}C{0.8cm}C{0.8cm}C{0.8cm}C{0.8cm}C{0.9cm}|C{0.9cm}}
\specialrule{.1em}{.05em}{.05em} 
   Body Part  & ~\cite{shotton2013real}  &~\cite{yub2015random} & ~\cite{jung2016sequential} & ~\cite{carreira2016human} &~\cite{haque2016towards} & Ours & Ours$^*$\\ \hline
    Head &  63.8  & 97.8 & 97.9 & 96.2 & 98.1 &  \textbf{98.7} & 99.7\\
    Neck &  86.4  & 95.8 & 93.5 & 85.2 & 97.5 &   \textbf{98.9} & 99.9\\
    Shoulders &  83.3  & 94.1 & 75.4 & 77.2 & \textbf{96.5} &  93.6 &  98.8\\
    Elbows &  73.2   & \textbf{77.9} & 41.1 & 45.4 &  73.3 &  69.2  & 88.6\\
    Hands &  51.3   & \textbf{70.5} & 19.9 & 30.9 & 68.7 &  59.8  & 82.9\\ \hhline{--------}
    Torso &  65.0   & 93.8 & 95.4 & 84.7 & 85.6 &   \textbf{97.9}  & 99.7\\
   Hips  &  50.8   & 80.3 & 74.6 & 83.5 & 72.0 &  \textbf{84.5}  & 94.8\\
   Knees  &  65.7   & 68.8 & 79.7 & 81.8 & 69.0 &  \textbf{87.3}  & 97.0\\
    Feet &  61.3   & 68.4 & 80.5 & 80.9 & 60.8 & \textbf{83.2}  & 88.0\\ \hhline{========}
    Upper Body &  70.7   & \textbf{84.8} & 58.0 & 61.0 &  84.0 & 80.3 & 92.5 \\ 
  Lower Body   &  59.3   & 72.5 & 80.7 & 82.1 & 67.3 &  \textbf{86.8} & 94.2\\
     Full Body&  65.8  & {\color{blue} \textbf{80.5}} & 68.6 & 71.0 & 77.4 &   {\color{red} \textbf{83.4}} &  93.3\\ \specialrule{.1em}{.05em}{.05em} 
\end{tabular}
\caption{Comparison with existing methods on the ITOP front-view dataset. {\it Ours$^*$} is the performance of the proposed method trained and tested from the 2D ground truth positions. Proposed model outperforms the existing methods.}
\label{table:comparison_ITOP_STOA}
\end{table}

We also investigate the relationship between the $xy$-size of the local patch and the performance, and the result is shown in Figure~\ref{fig:localSz}. As Figure~\ref{fig:localSz} shows, a  larger local patch can capture the global context more effectively so that our network can localize the body joint more accurately. However, as the receptive field of our model is 35$\times$35$\times$35, inputs larger than that size cannot fully provide additional contextual information, thereby resulting in a limited improvement in performance when the size is enlarged from 32$\times$32 to 40$\times$40. We cannot build a model whose receptive field size and input are larger than the proposed architecture because of  GPU memory shortage.

\begin{figure}[t]
\begin{center}
   \includegraphics[width=1.0\linewidth]{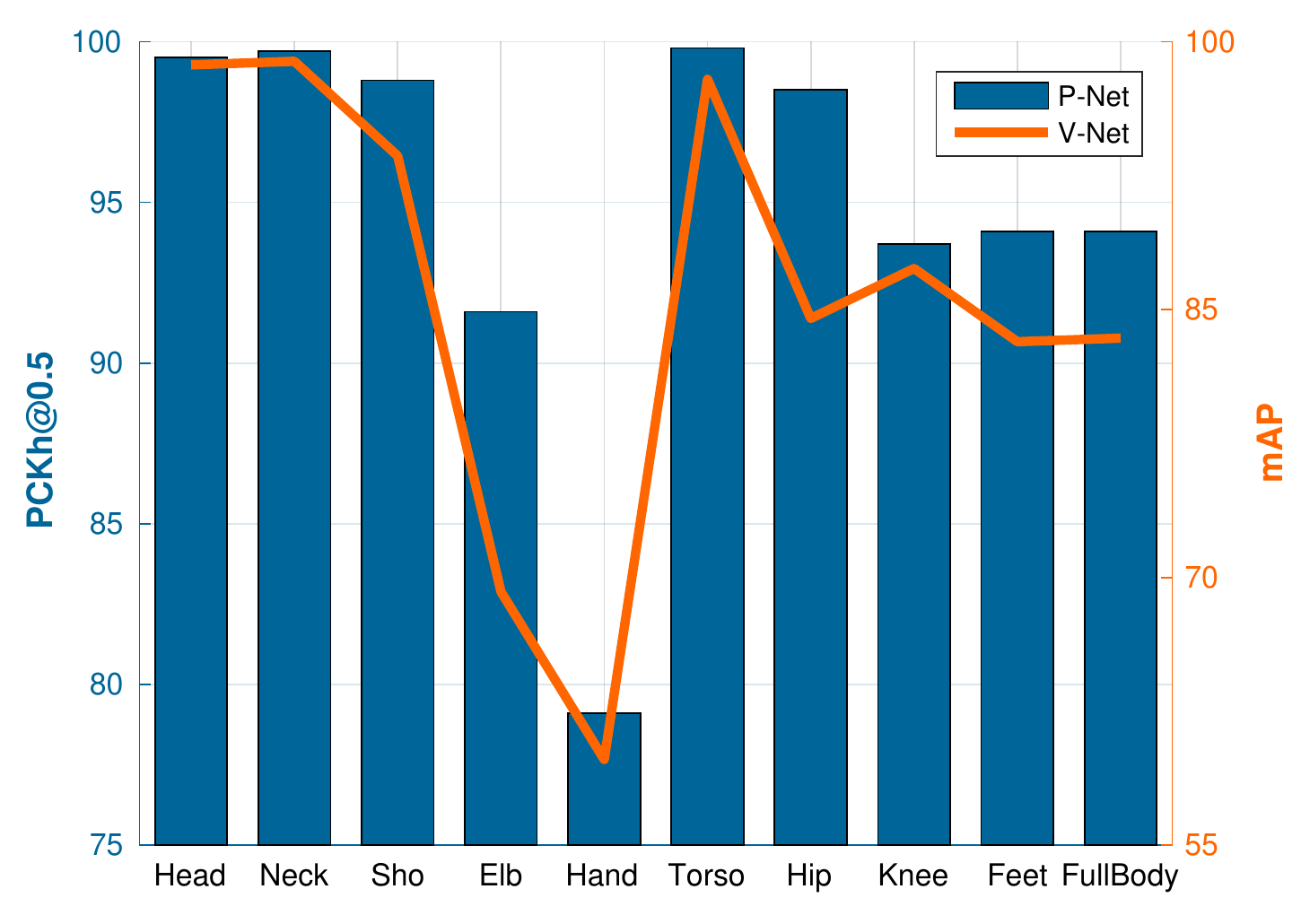}
\end{center}
   \caption{Performance of the P-Net and V-Net on the ITOP front-view dataset. The left axis represents PCKh@0.5 of the P-Net, and the right axis represents mAP from the 10 cm rule.}
\label{fig:pnet_vnet_comparison}
\end{figure}

\begin{figure*}
\begin{center}
\includegraphics[width=1.0\linewidth]{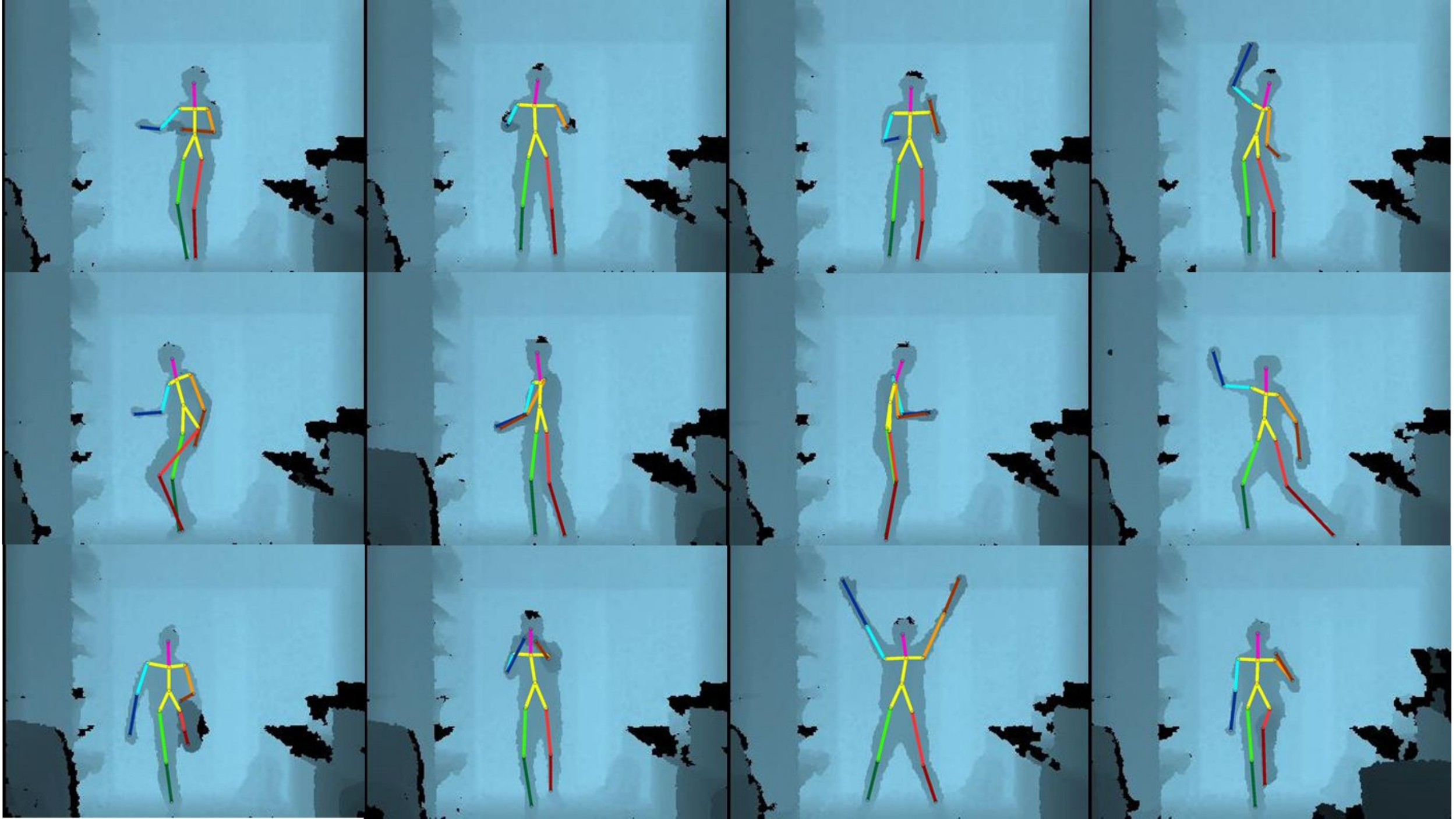}
\end{center}
   \caption{Qualitative results of our algorithm on the ITOP dataset}
\label{fig:itop_qualitative}
\end{figure*}

\subsection{Comparison with state-of-the-art methods}
We compared the proposed method with other state-of-the art methods, and results are given in Table~\ref{table:comparison_EVAL_STOA} and Table~\ref{table:comparison_ITOP_STOA}. The results of ~\cite{yub2015random,shotton2013real,carreira2016human} are obtained from ~\cite{haque2016towards}. We cannot evaluate the method in ~\cite{jung2016sequential} on the EVAL dataset because they need foreground and background labels, which are not provided in the EVAL dataset. Also, to evaluate ~\cite{jung2016sequential} on the ITOP dataset, we corrected some mislabeled foreground and background labels provided in the dataset by ourselves using the ground truth joint positions. Note that the corrected ground truth labels of foreground and background are used only for the evaluation of ~\cite{jung2016sequential} on the ITOP dataset. As the results show, our method outperforms the existing methods by a large margin. Our method  notably outperforms existing methods for both datasets. Compared with ~\cite{haque2016towards}, our method does not use the iterative refinement strategy, although their method used 10 steps of refinement. Also, our method takes 0.28 second per frame, while Haque \etal~\cite{haque2016towards} takes 1.7 second and Jung \etal~\cite{yub2015random} takes 0.1 second per frame.

\subsection{Performance analysis}
Although our proposed model outperforms existing methods in full body, it gives lower accuracy of elbows and hands which are considered as the hardest body joints to localize. The first reason is the inaccurate estimation from the P-Net for those body joints. As the last column of Table~\ref{table:comparison_EVAL_STOA} and Table~\ref{table:comparison_ITOP_STOA} show, when the V-Net is trained and tested based on the 2D ground truth, it outperforms the proposed method by 14.7 and 9.9 point for the full body joints of the EVAL and ITOP dataset, respectively. Especially, the accuracy of the elbows and hands is increased about 20 point for both datasets, which is relatively higher increase than other body joints. This means that the low performance in elbows and hands is due to the inaccurate estimation from the P-Net as Figure~\ref{fig:pnet_vnet_comparison} shows. This shows us that if our P-Net becomes more accurate, there is room to improve accuracy of our system. The second reason is the lack of the global context information in the second stage of the proposed system. As Table~\ref{table:comparison_EVAL_STOA} and Table~\ref{table:comparison_ITOP_STOA} show, even though the V-Net is trained and tested from the 2D ground truth positions, the accuracy of elbows and hands is relatively lower than the other body joints. Also, as Figure~\ref{fig:pnet_vnet_comparison} shows, there is much difference between the performance of the P-Net and V-Net in elbows although knees and feet in which the P-Net gives the similar performance with the elbows have small difference. Considering elbows and hands can be occluded more severely than other body joints, global context information can help the system localize those body joints~\cite{wei2016convolutional,newell2016stacked}. Although we tried to enlarge the size of local patch to capture more global context information as shown in Figure~\ref{fig:localSz}, the second stage only exploits the part of the human body, which still lacks of understanding of entire human body structure. Embedding the holistic view of human body in the second stage can help network obtain the global context information so that it can localize body joints more accurately. The qualitative results are shown in Figure~\ref{fig:itop_qualitative} for the ITOP dataset.

\section{Conclusion}
We propose a novel depth-based 3D human pose estimation approach. To overcome the limitation of the traditional coordinate regression methods using the highly non-linear mapping, our model represents the depth map as the occupancy grid model and estimates the per-voxel likelihood to localize body joint. Also, our two-stage approach enables our network to deal with increased computational cost caused by 3D representations. Our model outperforms existing methods by a large margin in publicly available datasets. We aim to apply the proposed method to 3D hand pose estimation, which is a more complex problem because of numerous occlusions. Embedding more contextual information in the second stage is another topic for future work.

\clearpage

\begin{titlepage}
   \vspace*{\stretch{1.0}}
   \begin{center}
      \Large\textbf{Supplementary Material for ``Holistic Planimetric prediction to Local Volumetric prediction for 3D Human Pose Estimation"}\\
   \end{center}
   \vspace*{\stretch{2.0}}
\end{titlepage}

In this supplementary material, we present more experimental results which could not be included in the main manuscript due to the lack of space.

\section{Accuracy Comparison of the Results from the P-Net and V-Net on the EVAL dataset}
\begin{figure}[h]
\begin{center}
   \includegraphics[width=1.0\linewidth]{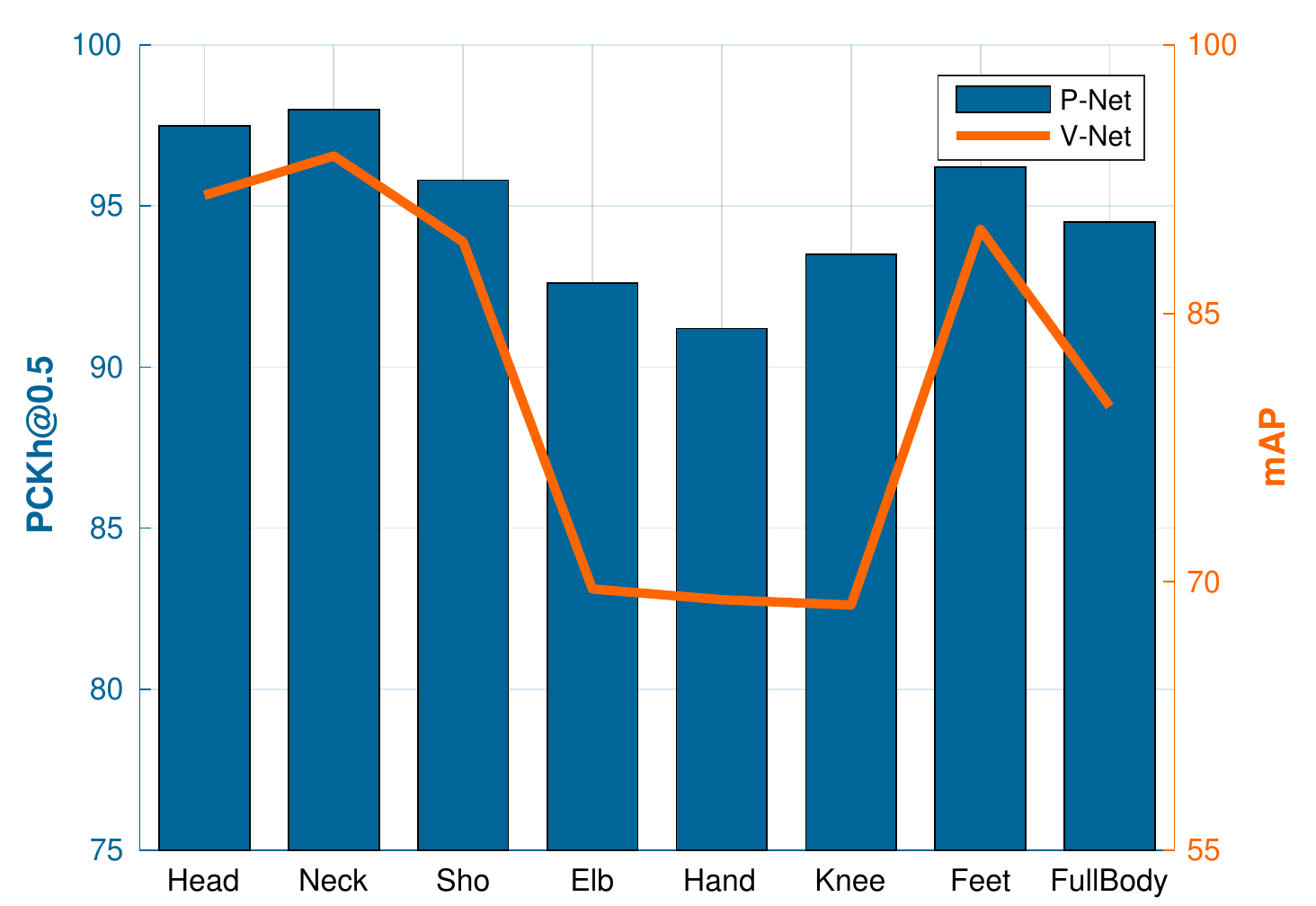}
\end{center}
   \caption{Performance of the P-Net and V-Net on the EVAL dataset. The left axis represents PCKh@0.5 of the P-Net, and the right axis represents mAP from the 10 cm rule.}
\label{fig:pnet_vnet_compare_EVAL}
\end{figure}

In Section 7.5 of the main manuscript, we analyzed the reason why the accuracy on elbows and hands are relatively lower than those of other body parts. 
Here, we provide more observations from the experiments on the EVAL dataset that supports our analysis.
Accuracy comparison of the results from the P-Net and V-Net on the EVAL dataset is shown in Figure~\ref{fig:pnet_vnet_compare_EVAL}. Firstly, the results from the P-Net gives relatively low accuracy on elbows, hands and knees, thus limiting the accuracy on corresponding parts in the V-Net. Secondly, the accuracy gap between the P-Net and V-Net for elbows, hands and knees are relatively larger than those for other body joints, due to the lack of global context information in the second stage. The trend is similar to the one in the ITOP dataset (Figure 7 in the main manuscript).

\section{Qualitative results}
\subsection{EVAL dataset}
\begin{figure*}
\begin{center}
\includegraphics[width=1.0\linewidth]{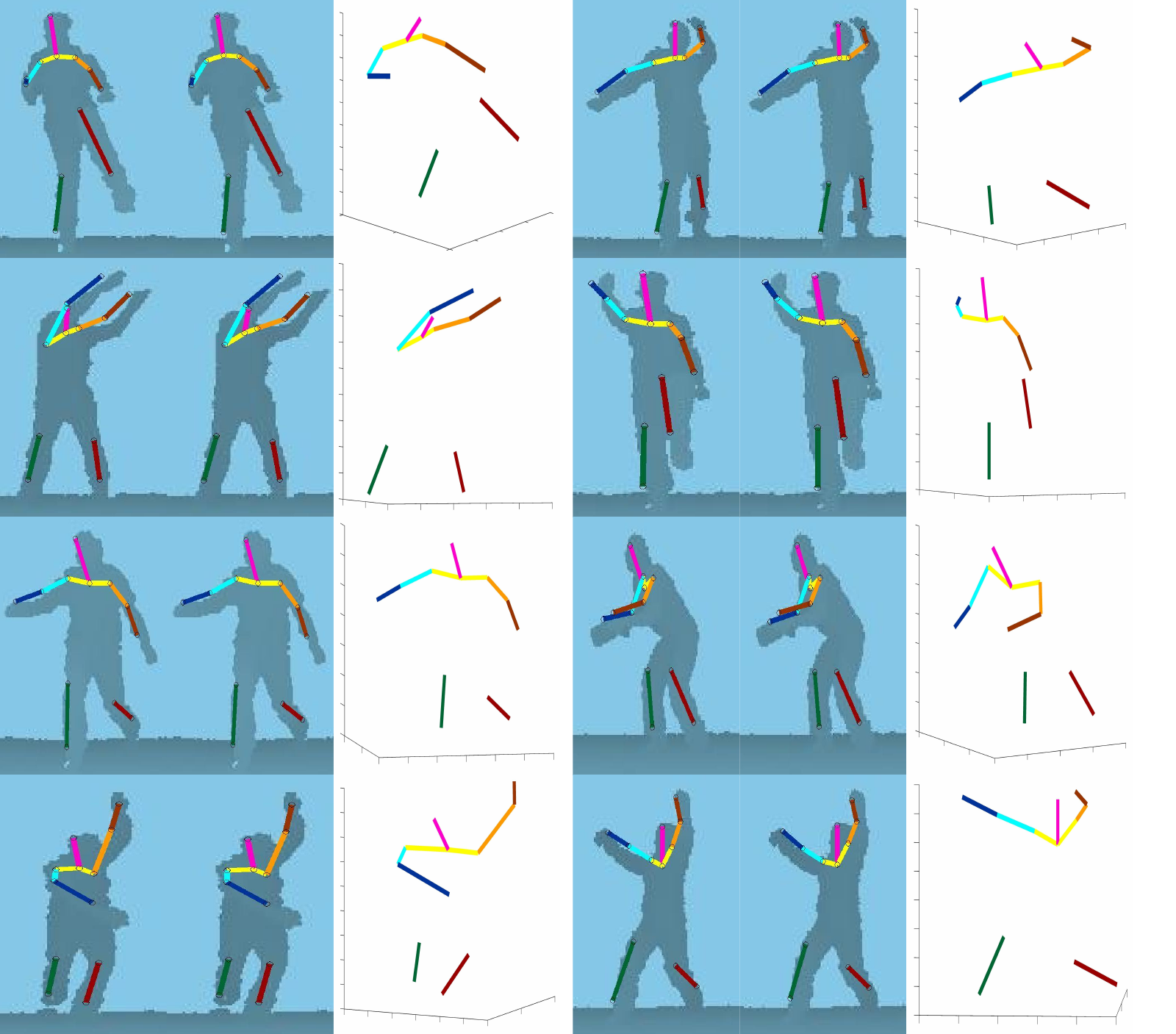}
\end{center}
   \caption{The estimation on the EVAL dataset. {\it Left: }Estimation of the P-Net. {\it Middle: }Estimation of the V-Net projected to the 2D space. {\it Right: }Estimation of the V-Net in the 3D space.}
\label{fig:example_eval}
\end{figure*}

Some example results of our method on the EVAL dataset are shown in Figure~\ref{fig:example_eval}.

\subsection{ITOP dataset}
\begin{figure*}
\begin{center}
\includegraphics[width=1.0\linewidth]{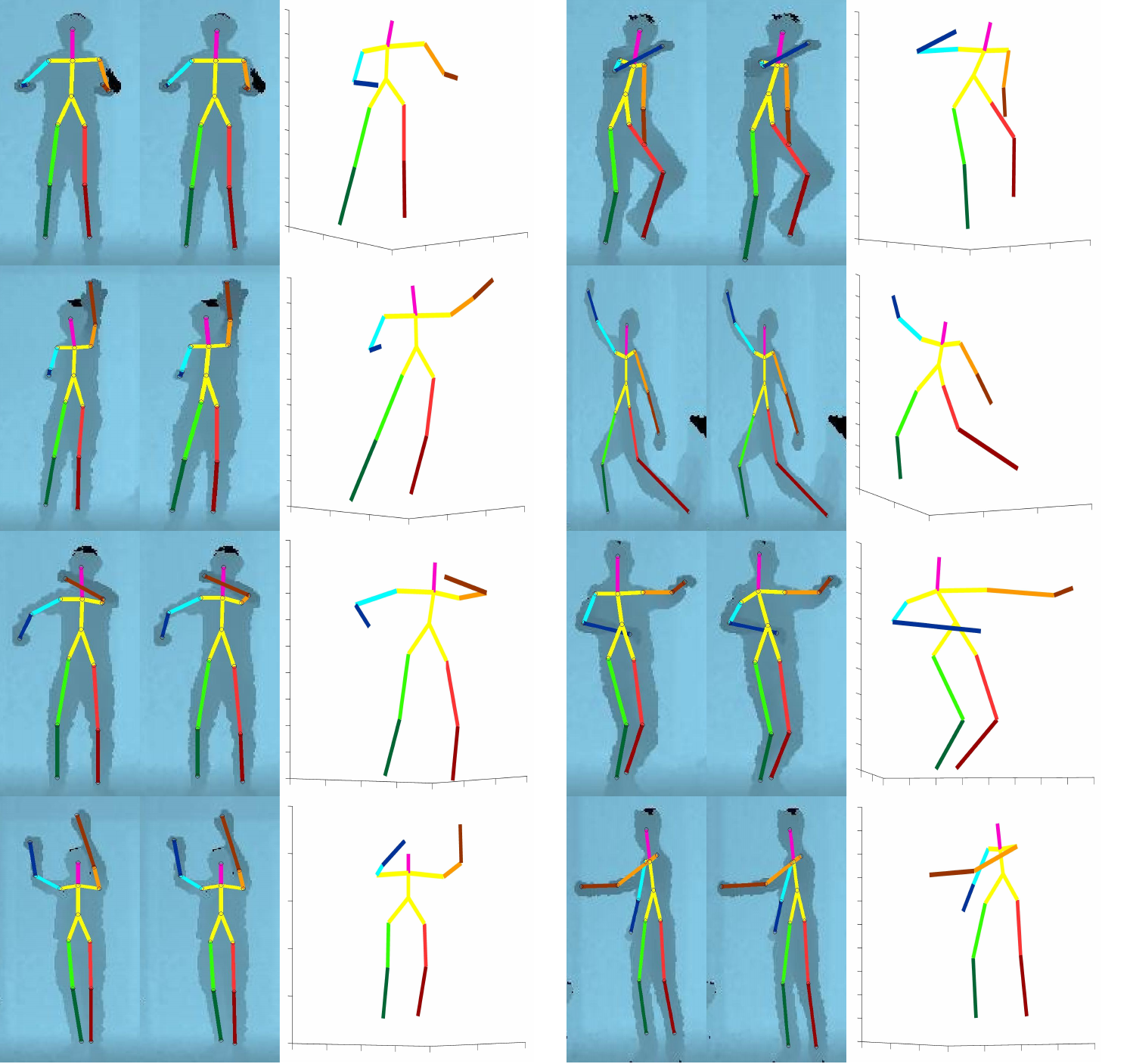}
\end{center}
   \caption{The estimation on the ITOP dataset. {\it Left: }Estimation of the P-Net. {\it Middle: }Estimation of the V-Net projected to the 2D space. {\it Right: }Estimation of the V-Net in the 3D space}
\label{fig:example_itop}
\end{figure*}

Some example results of our method on the ITOP dataset are shown in Figure~\ref{fig:example_itop}.


\end{document}